\documentclass[10pt,twocolumn,letterpaper]{article}

\usepackage{cvpr}
\usepackage{times}
\usepackage{epsfig}
\usepackage{graphicx}
\usepackage{amsmath}
\usepackage{amssymb}

\usepackage{algorithm}
\usepackage{algpseudocode}

\usepackage{array}
\usepackage{tabularx}
\usepackage[table,xcdraw]{xcolor}

\usepackage{colortbl}
\makeatletter
\@namedef{ver@everyshi.sty}{}
\makeatother
\usepackage{pgfplots}
\usepackage{pgfplotstable}

\usepackage[breaklinks=true,bookmarks=false]{hyperref}

\cvprfinalcopy 

\def\cvprPaperID{1962} 

\begin{document}

\title{Towards Visual Feature Translation}

\author{
    Jie Hu$^1$, Rongrong Ji$^{12}$\thanks{Corresponding author.}, Hong Liu$^1$, Shengchuan Zhang$^1$, Cheng Deng$^3$, and Qi Tian$^4$\\
    $^1$Fujian Key Laboratory of Sensing and Computing for Smart City, Department of Cognitive Science,\\
    School of Information Science and Engineering, Xiamen University, Xiamen, China.\\
    $^2$Peng Cheng Laboratory, Shenzhen, China. $^3$Xidian University. $^4$Noah'’s Ark Lab, Huawei.\\
    {\tt\small \{hujie.cpp,lynnliu.xmu,chdeng.xd\}@gmail.com, \{rrji,zsc\_2016\}@xmu.edu.cn, tian.qi1@huawei.com}
}
\pagestyle{empty}  
\maketitle

\begin{abstract}
Most existing visual search systems are deployed based upon fixed kinds of visual features, which prohibits the feature reusing across different systems or when upgrading systems with a new type of feature.
Such a setting is obviously inflexible and time/memory consuming, which is indeed mendable if visual features can be ``translated" across systems.
In this paper, we make the first attempt towards visual feature translation to break through the barrier of using features across different visual search systems.
To this end, we propose a Hybrid Auto-Encoder (HAE) to translate visual features, which learns a mapping by minimizing the translation and reconstruction errors.
Based upon HAE, an Undirected Affinity Measurement (UAM) is further designed to quantify the affinity among different types of visual features.
Extensive experiments have been conducted on several public datasets with sixteen different types of widely-used features in visual search systems.
Quantitative results show the encouraging possibilities of feature translation.
For the first time, the affinity among widely-used features like SIFT and DELF is reported.
\end{abstract}

\thispagestyle{empty} 
\section{Introduction}
\begin{figure}[!t]
\centering
\includegraphics[width=3.0in]{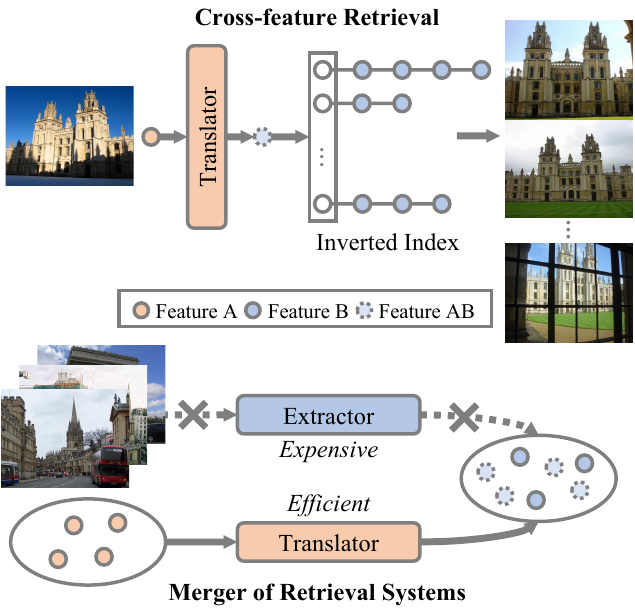}
\caption{
Two potential applications of visual feature translation.
Top: In cross-feature retrieval, Feature A is translated to Feature AB, which can be used to search images that are represented and indexed by Feature B.
Bottom: In the merger of retrieval systems, Feature A used in System A is efficiently translated to Feature AB, instead of the expensive process of re-extracting entire dataset in System A with Feature B.
}
\label{fig1}
\end{figure}

Visual features serve as the basis for most existing visual search systems.
In a typical setting, a visual search system can only handle pre-defined features extracted from the image set offline.
Such a setting prohibits the reusing of a certain kind of visual feature across different systems.
Moreover, when upgrading a visual search system, a time-consuming step is needed to extract new features and to build the corresponding indexing, while the previous features and indexing are simply discarded.
Breaking through such a setting, if possible, is by any means very beneficial.
For instance, the existing features and indexing can be efficiently reused when updating old features with new ones, which can significantly save the time and memory cost.
For another instance, images can be efficiently archived with only respective features for cross-system retrieval.
These examples are detailedly depicted in Fig.~\ref{fig1}.
However, feature reusing is not an easy task.
Various dimensions and diverse distributions of different types of features prohibit reusing features directly.
Therefore, a feature ``translator" is needed to transform across different types of features, which, to our best knowledge, remains untouched in the literature.
Intuitively, given a set of images extracted with different types of features, one can leverage the feature pairs to learn the corresponding feature translator.
\begin{figure*}[!t]
\centering
\includegraphics[width=6.0in]{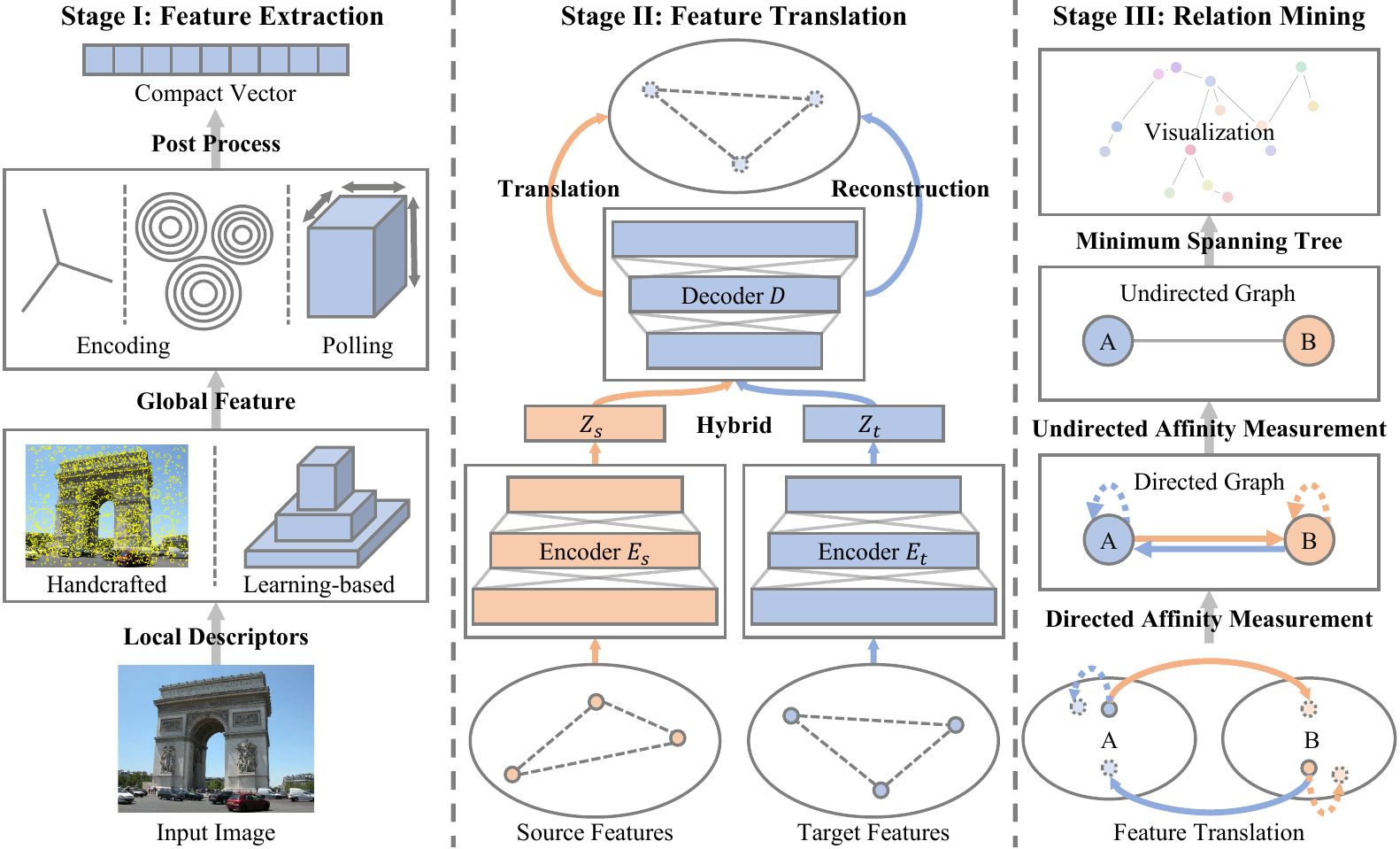}
\caption{
The overall flowchart of the proposed visual feature translation.
In Stage I, different handcrafted or learning-based features are extracted from image set for training.
In Stage II, the mappings from source features to target features are learned by our HAE with the encoders $E_s, E_t$ and the decoder $D$.
Then the encoder $E_s$ and the decoder $D$ are used in inference.
In Stage III, the UAM is calculated to quantify the affinity among different types of visual features, which is further visualized by employing the Minimum Spanning Tree. 
}
\label{fig2}
\end{figure*}
In this paper, we make the first attempt to investigate \emph{visual feature translation}.
Concretely, we propose a Hybrid Auto-Encoder (HAE) that learns a mapping from source features to target features by minimizing the translation and reconstruction errors.
HAE consists of two encoders and one decoder.
In training, the source and target features are encoded into a latent space by corresponding encoders.
Features in this latent space are sent to a shared decoder to produce the translated features and reconstructed features.
Then the reconstruction and translation errors are minimized by optimizing the objective function.
In inference, the encoder of source features and the shared decoder are used for translation.
The proposed HAE further provides a way to characterize the affinity among different types of visual features.
Based upon HAE, an Undirected Affinity Measurement (UAM) is further proposed, which provides, also for the first time, a quantification of the affinity among different types of visual features.
We also discover that UAM can predict the translation quality before the actual translation happens.
We train HAE on the Google-Landmarks dataset \cite{hyeonwoo2017large} and evaluate in total $16$ different types of widely-used features in visual search community \cite{arandjelovic2012three, babenko2015aggregating, Jegou2010hf, kalantidis2016cross, lowe2004distinctive, perronnin2010large, radenovic2018fine, razavian2016visual, tolias2015particular}.
The tests of feature translation are conducted on three benchmark datasets, \ie, Oxford5k \cite{radenovic2018revisiting}, Paris6k \cite{philbin2007object}, and Holidays \cite{jegou2008hamming}.
Quantitative results show the encouraging possibility for feature translation.
In particular, HAE works relatively well for feature pairs such as V-CroW to V-SPoC (\eg, 0.1 mAP decrease on the Oxford5k benchmark) and R-rMAC to R-CroW (\eg, 1.8 mAP decrease on the Holidays benchmark).
Interestingly, visual feature translation provides some intriguing results (see Fig.~\ref{fig4} later in our experiments).
For example, when translating from SIFT to DELF, characteristics like rotation or viewpoint invariance can be highlighted, which provides a new way to absorb merits of handcrafted features to learning-based ones.
In short, our contributions can be summarized as below:
\begin{itemize}
    \vspace{-6px}
    \item We are the first to address the problem of visual feature translation, which fills in the gaps between different types of features.
    \vspace{-6px}
    \item We are the first to quantify the affinity among different types of visual features in retrieval, which can be used to predict the quality of feature translation.
    \vspace{-6px}
    \item The proposed scheme innovates in several detailed designs, such as the HAE for training the translator and the UAM for quantifying the affinity. The source code and meta-data are released online\footnote{https://github.com/hujiecpp/VisualFeatureTranslation}.
    \vspace{-6px}
\end{itemize}
The rest of this paper is organized as follows.
Section \ref{sec:RW} reviews the related work.
The proposed feature translation and feature relation mining algorithms are introduced in Section \ref{sec:FT}.
Quantitative experiments are given in Section \ref{sec:experiments}.
Finally, we conclude this work in Section \ref{sec:conclusion}.
\section{Related Work}
\label{sec:RW}
\textbf{Visual Feature.}
Early endeavors mainly include holistic features (\eg, color histogram \cite{huang1997image} and shape \cite{belongie2002shape}) and handcrafted local descriptors \cite{bay2006surf, jegou2010accurate, perdoch2009efficient, mikolajczyk2005comparison, nister2006scalable, qin2011hello, simonyan2014learning,sivic2003video}, such as SIFT \cite{lowe2004distinctive} and ORB \cite{rublee2011orb}.
Then, different aggregation schemes (\eg, Fisher Vector \cite{perronnin2010large} and VLAD \cite{Jegou2010hf}) are proposed to encode local descriptors.
Along with the proliferation of neural networks, deep visual features have dominated visual search \cite{arandjelovic2016netvlad, babenko2015aggregating,babenko2014neural,gordo2016deep, hyeonwoo2017large,kalantidis2016cross, ng2015exploiting, radenovic2018fine, razavian2014cnn,tolias2015particular}, for instance, the local feature DELF \cite{hyeonwoo2017large} and the global feature produced by GeM \cite{radenovic2018fine} pooling are both prominent for representing images.
Detailed surveys of visual features can be found in \cite{smeulders2000content, zheng2017SIFT}.

\textbf{Transfer Learning.}
Transfer learning \cite{pan2010survey, tan2018survey} aims to improve the learning of the target task using the knowledge in source domain.
It can be subdivided into: instance transfer, feature transfer, parameter transfer, and relation transfer.
Our work relates to, but is not identical with, the feature transfer.
Feature transfer \cite{argyriou2007multi, dai2007co, duan2012domain, gretton2012optimal, liu2011cross, long2014transfer, pan2011domain, raina2007self, tzeng2014deep} is usually based on the hypothesis that the source domain and target domain have some shared characteristics.
It aims to find a common feature space for both source and target domains, which serves as a new representation to \emph{improve the learning of the target task}.
For instance, the Structural Corresponding Learning \cite{blitzer2006domain} uses pivot features to learn a mapping from features of both domains to a shared feature space.
For another instance, Joint Geometrical and Statistical Alignment \cite{zhang2017joint} learns two coupled projections that project features of both domains into subspaces where the geometrical and distribution shifts are reduced.
More recently, deep learning has been introduced into feature transfer \cite{long2015learning, long2016deep2, long2016deep, sener2016learning}, in which neural networks are used to find the common feature spaces.
In contrast, the visual feature translation aims to learn a mapping to translate features from the source space to the target space, and the translated features are \emph{used directly in the target space}.
\section{Visual Feature Translation}
\label{sec:FT}
Fig.~\ref{fig2} shows the overall flowchart of the proposed visual feature translation.
Firstly, source and target feature pairs are extracted from image set for training in Stage I.
Then, feature translation based on HAE is learned in Stage II.
After translation, the affinity among different types of features is quantified and visualized in Stage III.
\subsection{Preprocessing}
As shown in Stage I of Fig.~\ref{fig2}, we prepare the source and target features for training the subsequent translator.
For the handcrafted features such as SIFT \cite{lowe2004distinctive}, the local descriptors are extracted by the designed procedures firstly.
These local descriptors are then aggregated by encoding schemes to produce the global features.
For the learning-based features such as V-MAC \cite{razavian2016visual, tolias2015particular}, the feature maps are extracted by neural networks firstly, followed by a pooling layer or encoding schemes to produce the feature vectors.
In our settings, we investigate in total $16$ different types of features, a detailed table~of which can be found in Table~\ref{table1}.
The feature sets are arranged to form $16\times16$ feature set pairs $(\mathcal{V}_s, \mathcal{V}_t)$, where $\mathcal{V}_s$ denotes the set of source features and $\mathcal{V}_t$ denotes the set of target features.
The implementation is detailed in Section \ref{sec:implementation}.
\subsection{Learning to Translate}
\label{sec:stage2}
To achieve the task of translating different types of features, a Hybrid Auto-Encoder (HAE) is proposed, which is shown in Stage II of Fig.~\ref{fig2}.
For training HAE, the source features $\mathcal{V}_s$ and the target features $\mathcal{V}_t$ are input to the model which outputs the translated features $\mathcal{V}_{st}$ and the reconstructed features $\mathcal{V}_{tt}$.
%
%
%

%
\begin{algorithm}[t]
\caption{The Training of HAE}
\label{alg1}
\hspace*{0.02in} {\bf Input:} %
Feature sets $\mathcal{V}_s$ and $\mathcal{V}_t$, decoders $E_s, E_t$ and encoder $D$ parameterized by $\theta_{E_s},\theta_{E_t}$ and $\theta_{D}$.\\
\hspace*{0.02in} {\bf Output:} %
The learned translator $E_s$ and $D$.
\begin{algorithmic}[1]
\While{not convergence}
    \State Get $\mathcal{Z}_s$ by $\mathcal{Z}_s = E_s(\mathcal{V}_s)$.
    \State Get $\mathcal{Z}_t$ by $\mathcal{Z}_t = E_t(\mathcal{V}_t)$.
    \State Get $\mathcal{V}_{st}$ by translation: $\mathcal{V}_{st} = D(\mathcal{Z}_s)$.
    \State Get $\mathcal{V}_{tt}$ by reconstruction: $\mathcal{V}_{tt} = D(\mathcal{Z}_t)$.
    \State Optimize the Eq.~\ref{eq1}. 
\EndWhile
\State \Return $E_s$ and $D$.
\end{algorithmic}
\end{algorithm}
Formally, HAE consists of two encoders $E_s, E_t$ and one decoder $D$.
In training, $v_s \in \mathcal{V}_s$ is encoded into the latent feature $z_s \in \mathcal{Z}_s$ by the encoder $E_s$, and the same for $v_t \in \mathcal{V}_t$ into $z_t \in \mathcal{Z}_t$ by $E_t$.
The latent features $z_s$ and $z_t$ are then decoded to obtain the translated feature $v_{st} \in \mathcal{V}_{st}$ and the reconstructed feature $v_{tt} \in \mathcal{V}_{tt}$ by the shared decoder $D$.
We define the Euclidean distance as $\mathcal{E}(x,y)=\|x-y\|_2$.
The $E_s, E_t$ and $D$ are parameterized by $\theta_{E_s},\theta_{E_t}$ and $\theta_{D}$, which is learned by minimizing the following loss function:
\begin{equation}
\begin{split}
\label{eq1}
\mathcal{L}(\theta_{E_s}, \theta_{E_t}, \theta_{D}) = &\mathbb{E}_{v_{st} \in \mathcal{V}_{st}, v_t \in \mathcal{V}_t}[\mathcal{E}(v_{st}, v_t)] \\ +& \mathbb{E}_{v_{tt} \in \mathcal{V}_{tt}, v_t \in \mathcal{V}_t}[\mathcal{E}(v_{tt}, v_t)],
\end{split}
\end{equation}
where we define the first item as the translation error and the second item as the reconstruction error.
In the processing of the feature translation, only $E_s$ and $D$ are used to translate features from $\mathcal{V}_s$ to $\mathcal{V}_t$.
The algorithm for training the HAE is summarized as Alg.~\ref{alg1}.

We then get the following characteristics for our visual feature translation:
\noindent\textbf{Characteristic I: Saturation.} 
The performance of translated features is difficult to exceed that of the target features.
This phenomenon is inherent in the feature translation process.
According to Eq.~\ref{eq1}, the translation and reconstruction errors are minimized after optimizing.
However, they are difficult to approach zero due to the information loss brought by the architecture of Auto-Encoder.
\noindent\textbf{Characteristic II: Asymmetry.}
The convertibility of translation is discrepancy between A2B and B2A (We abbreviate A2B for the translation from features A to features B, \etc).
The networks for translating different types of features are by nature asymmetry.
HAE relies on the translation error and reconstruction error, which is not the same between A2B and B2A.
\noindent\textbf{Characteristic III: Homology}.
In general, homologous features tend to have high convertibility.
In contrast, the convertibility is not guaranteed for heterogenous features.
Homologous features refer to the features extracted by the same extractor but encoded or pooled by different methods (\eg, DELF-FV \cite{hyeonwoo2017large, perronnin2010large} and DELF-VLAD \cite{hyeonwoo2017large, Jegou2010hf}, or V-CroW \cite{kalantidis2016cross} and V-SPoC \cite{babenko2015aggregating}), and the heterogenous features refer to the features extracted by different extractor.
This characteristic is analyzed in details in Section \ref{sec:Translation_result}.
\subsection{Feature Relation Mining}
\label{sec:RM}
HAE provides a way to quantify the affinity between feature pairs.
Therefore, the affinity among different types of features can be quantified as the Stage III shown in Fig.~\ref{fig2}.
First, we use the difference between translation and reconstruction errors as a Directed Affinity Measurement (DAM) and calculate the directed affinity matrix $M$ which forms a directed graph for all feature pairs.
Second, in order to quantify the total affinity among features, we design an Undirected Affinity Measurement (UAM) by employing $M$.
The calculated undirected affinity matrix $U$ is symmetry, which forms a complete graph.
Third, we visualize the local similarity between features by using the Minimum Spanning Tree (MST) of the complete graph.
\begin{algorithm}[t]
\caption{Affinity Calculation and Visualization}
\label{alg2}
\hspace*{0.02in} {\bf Input:} %
The number of different types of features $n$, the feature pairs $(\mathcal{V}_s, \mathcal{V}_t)$ and the translator $E_s, D$.\\
\hspace*{0.02in} {\bf Output:} %
The directed affinity matrix $M$ and the undirected affinity matrix $U$.
\begin{algorithmic}[1]
\For{i = 1 : n, j = 1 : n}
　　\State Calculate $M_{i \to j}$ by Eq.~\ref{eq8}.
\EndFor
\For{i = 1 : n, j = 1 : n}
　　\State Calculate $R_{i \to j}$ and $C_{i \to j}$ by Eq.~\ref{eq9} and Eq.~\ref{eq10}.
\EndFor \\
Calculate $U$ by Eq.~\ref{eq11}.\\
Generate the MST based on $U$ by Kruskal'’s algorithm.\\
Visualize the MST.
\State \Return $M, U$.
\end{algorithmic}
\end{algorithm}

\begin{figure*}[!t]
\centering
\includegraphics[width=6.71in]{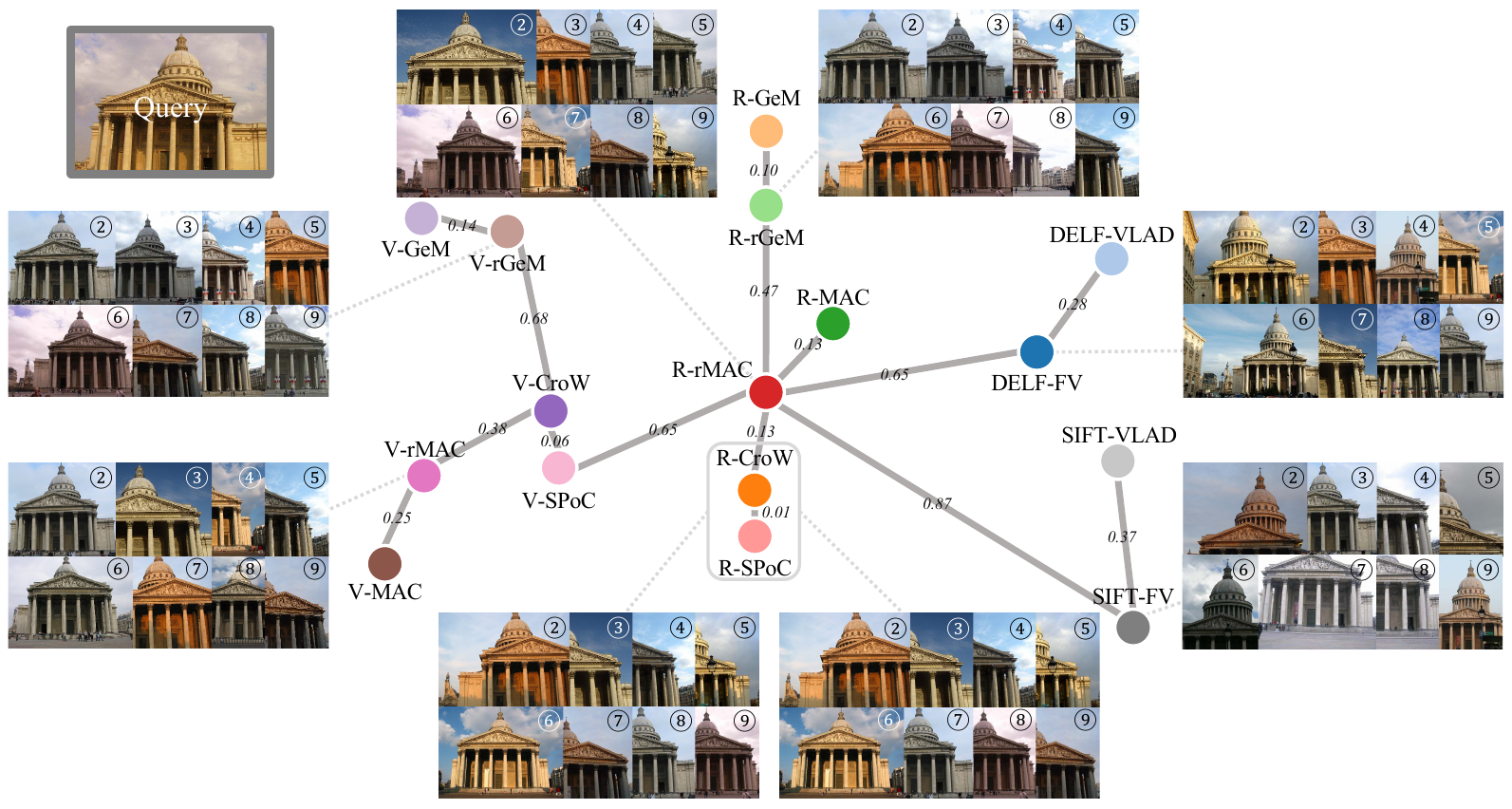}
\caption{
The visualization of the MST based on $U$ with popular visual search features.
The length of edges is the average value of the results on Holidays, Oxford5k and Paris6k datasets.
The images are the retrieval results for a query image of the Pantheon with corresponding features in the main trunk of the MST.
The close feature pairs such as R-SPoC and R-CroW have similar ranking lists.
}
\label{fig3}
\end{figure*}
\textbf{Directed Affinity Measurement.}
We assume that after optimizing, for Eq.~\ref{eq1}, the reconstruction error is smaller than the translation error.
This intuitive assumption is verified later in Section \ref{sec:Relation_result}.
Then, we can find that:
\begin{equation}
\begin{split}
\label{eq7}
    \mathcal{L} \geq &\mathbb{E}_{v_{st} \in \mathcal{V}_{st}, v_t \in \mathcal{V}_t}[\mathcal{E}(v_{st}, v_t)] \\ - &\mathbb{E}_{v_{tt} \in \mathcal{V}_{tt}, v_t \in \mathcal{V}_t}[\mathcal{E}(v_{tt}, v_t)] \geq 0.
\end{split}
\end{equation}
According to this inequation, when minimizing $\mathcal{L}$, the translation error is forced to approximate the reconstruction error.
If translation error is close to reconstruction error, we think the translation between source and target features is similar to the reconstruction of target features, which indicates the source and target features have high affinity.
Therefore, we regard the difference between the translation and reconstruction errors as the affinity measurement.
We use $M_{s \to t}$ to represent the DAM between $\mathcal{V}_s$ and $\mathcal{V}_t$.
The calculation of the element at row $s$ and column $t$ of $M$ is defined as follows:
\begin{equation}
\begin{split}
\label{eq8}
M_{s \to t} = &\mathbb{E}_{v_{st} \in \mathcal{V}_{st}, v_t \in \mathcal{V}_t}[\mathcal{E}(v_{st}, v_t)] \\ - &\mathbb{E}_{v_{tt} \in \mathcal{V}_{tt}, v_t \in \mathcal{V}_t}[\mathcal{E}(v_{tt}, v_t)].
\end{split}
\end{equation}
%
%

%
\textbf{Undirected Affinity Measurement.}
Due to the asymmetry characteristic, $M$ is asymmetric, which is unsuitable~to be the total affinity measurement of feature pairs.
We then resort to designing an Undirected Affinity Measurement (UAM) to quantify the overall affinity among different types of features.
Specifically, we treat A2B and B2A as a unified whole, therefore the rows and columns of $M$ are considered consistently.
For the rows of $M$, the element at row $i$ and column $j$ of the matrix $R$ with normalized rows is defined as:
\begin{equation}
\label{eq9}
R_{i\to j}= \frac{M_{i\to j} - \min(M_{i\to :})}{\max(M_{i\to :}) - \min(M_{i\to :})},
\end{equation}
where $\min(M_{i\to :})$ and $\max(M_{i\to :})$ are the minimum and maximum of the row $i$, and $R_{i \to j}$ is normalized to $[0, 1]$.
In a similar way, for the columns of $M$, the element at row $i$ and column $j$ of the matrix $C$ with normalized columns is defined as:
\begin{equation}
\label{eq10}
C_{i\to j}= \frac{M_{i\to j} - \min(M_{:\to j})}{\max(M_{:\to j}) - \min(M_{:\to j})},
\end{equation}
where $\min(M_{:\to j})$ and $\max(M_{:\to j})$ are the minimum and maximum of the column $j$, and $C_{i \to j}$ is normalized to $[0, 1]$.
The undirected affinity matrix $U$ is defined as follows:
\begin{equation}
\label{eq11}
U = \frac{1}{4}(R+R^T+C+C^T).
\end{equation}
If $U_{ij}$ has a small value, feature $i$ and feature $j$ are similar, and vice versa.
\textbf{The Visualization.}
We use the Minimum Spanning Tree (MST) to visualize the relationship of features based on $U$.
The Kruskal's algorithm \cite{kruskal1956shortest} is used to find MST.
This algorithm firstly creates a forest $G$, where each vertex is a separate tree.
Then the edge with minimum weight that connects two different trees is recurrently added to the forest $G$, which combines two trees into a single tree.
The final output forms an MST for the complete graph.
The MST helps us to understand the most related feature pairs (connected by an edge), as well as their affinity score (the length of the edge).
The overall procedure is summarized as Alg.~\ref{alg2}.
The visualization result of the affinity among popular visual features with a query example can be found in Fig.~\ref{fig3}.
\section{Experiments}
\label{sec:experiments}
We show the experiments in this section.
First, we introduce the experimental settings.
Then, the translation performance of our HAE is reported.
Finally, we visualize and analyze the results of relation mining.
\subsection{Experimental Settings}
\label{sec:implementation}
\textbf{Training Dataset.}
The Google-Landmarks dataset \cite{hyeonwoo2017large} contains more than 1M images captured at various landmarks all over the world.
We randomly pick 40,000 images from this dataset to train HAE, and pick 4,000 other images to train PCA whitening \cite{babenko2015aggregating, jegou2012negative} and creating the codebooks for local descriptors.
\textbf{Test Dataset.}
We use the Holidays, Oxford5k and Paris6k datasets for testing. 
The Holidays dataset \cite{jegou2008hamming} has 1,491 images with various scene types and 500 query images.
The Oxford5k dataset \cite{philbin2007object} consists of 5,062 images which have been manually annotated to generate a comprehensive ground truth for 55 query images.
Similarly,  the Paris6k dataset \cite{philbin2008lost} consists of 6,412 images with 55 query images.
Since the scalability of retrieval algorithms is not our main concern, we do not use the disturbance dataset Flickr100k \cite{philbin2008lost}.
Recently, the work in \cite{radenovic2018revisiting} revisited the labels and queries on both Oxford5k and Paris6k.
Because the images remained the same, which does not affect the characteristics of features, we do not use the revisited datasets as our test datasets.
The mean average precision (mAP) is used to evaluate the retrieval performance.
We translate the source features of reference images to the target space, and the target features of query images are used for testing.
\textbf{Features.}
L1 normalization and square root \cite{arandjelovic2012three} are applied to SIFT \cite{lowe2004distinctive}.
The original extraction approach (at most 1,000 local representations per image) is applied to DELF \cite{hyeonwoo2017large}.
%
%
The codebooks of FV \cite{perronnin2010large} and VLAD \cite{Jegou2010hf} are created for SIFT and DELF.
We use 32 components of Gaussian Mixture Model (GMM) to form the codebooks of FV and the dimension of this feature is reduced to 2,048 by PCA whitening.
The aggregated features are termed as SIFT-FV and DELF-FV.
We use 64 central points to form the codebooks of VLAD and the dimension of this feature is also reduced to 2,048 by PCA whitening.
The aggregated features are termed as SIFT-VLAD and DELF-VLAD.
For off-the-shelf deep features, we use ImageNet \cite{deng2009imagenet} pre-trained VGG-16 (abbreviated as V) \cite{simonyan2014very} and ResNet101 (abbreviated as R) \cite{he2016deep} to produce the feature maps.
The max-pooling (MAC) \cite{razavian2016visual, tolias2015particular}, average-pooling (SPoC) \cite{babenko2015aggregating}, weighted sum-pooling (CroW) \cite{kalantidis2016cross}, and regional max-pooling (rMAC) \cite{tolias2015particular} are then used to pool the feature maps.
The extracted features are termed as V-MAC, V-SPoC, V-CroW, V-rMAC, R-MAC, R-SPoC, R-CroW and R-rMAC, respectively.
For fine-tuned deep features, we consider the generalized mean-pooling (GeM) and regional generalized mean-pooling (rGeM) \cite{radenovic2018fine}.
The extracted features are termed as V-GeM, V-rGeM, R-GeM and R-rGeM, respectively.
\textbf{Network Architecture.} The task-specific network architectures of HAE have a fixed latent feature space of 510 dimension.
The parameter settings of encoder which consists of fully-connect layers with ReLU-based activation function are 2048-2048-2048-510 or 512-512-510 for encoding the features with 2048 or 512 dimension.
The parameter settings of the decoder are in reverse of that of encoder, depending on the dimension of the output features.
The output features are L2 normalized.
We use Multi-Layer Perceptron (MLP) as our baseline, whose architecture are 2048-2048-2048 or 512-512-512 for encoding the features with 2048 or 512 dimension, and the encoders are in reverse.
We used Adam \cite{kingma2014adam} optimizer to minimize the objective function for all feature pairs, where the learning rate is set as 0.00001.
\begin{table}
\begin{center}
\scalebox{0.8}[0.8]{
\begin{tabular}{c|c|c|c}
\hline
 & \textbf{Holidays} & \textbf{Oxford5k} & \textbf{Paris6k} \\ \hline
DELF-FV \cite{hyeonwoo2017large,perronnin2010large}   & 83.42             & 73.38             & 83.06            \\ \hline
DELF-VLAD \cite{hyeonwoo2017large,Jegou2010hf} & 84.61             & 75.31             & 82.54            \\ \hline
R-CroW \cite{kalantidis2016cross}    & 86.38             & 61.73             & 75.46            \\ \hline
R-GeM \cite{radenovic2018fine}     & 89.08             & 84.47             & 91.87            \\ \hline
R-MAC \cite{razavian2016visual, tolias2015particular}     & 88.53             & 60.82             & 77.74            \\ \hline
R-rGeM \cite{radenovic2018fine}    & 89.32             & 84.60             & 91.90            \\ \hline
R-rMAC \cite{tolias2015particular}    & 89.08             & 68.46             & 83.00            \\ \hline
R-SPoC \cite{babenko2015aggregating}    & 86.57             & 62.36             & 76.75            \\ \hline
V-CroW \cite{kalantidis2016cross}    & 83.17             & 68.38             & 79.79            \\ \hline
V-GeM \cite{radenovic2018fine}     & 84.57             & 82.71             & 86.85            \\ \hline
V-MAC \cite{razavian2016visual, tolias2015particular}     & 74.18             & 60.97             & 72.65            \\ \hline
V-rGeM \cite{radenovic2018fine}    & 85.06             & 82.30             & 87.33            \\ \hline
V-rMAC \cite{tolias2015particular}    & 83.50             & 70.84             & 83.54            \\ \hline
V-SPoC \cite{babenko2015aggregating}    & 83.38             & 66.43             & 78.47            \\ \hline
SIFT-FV \cite{arandjelovic2012three, lowe2004distinctive, perronnin2010large} & 61.77 & 36.25 & 36.91 \\ \hline
SIFT-VLAD \cite{arandjelovic2012three, lowe2004distinctive, Jegou2010hf} & 63.92 & 40.49 & 41.49 \\ \hline
\end{tabular}}
\end{center}
\caption{The mAP (\%) of target features.}
\label{table1}
\end{table}
\definecolor{a}{rgb}{0.6,0.8,0.5}
\def\ho#1{%
    \pgfmathsetmacro\calc{(#1-1.180)*100/(70.680-1.180)}
    \edef\clrmacro{\noexpand\cellcolor{a!\calc}}%
    \clrmacro%
    \textcolor{black}{#1}
}
\definecolor{b}{rgb}{0.7,0.8,0.9}
\def\ox#1{%
    \pgfmathsetmacro\calc{(#1-0.060)*100/(81.180-0.060)}
    \edef\clrmacro{\noexpand\cellcolor{b!\calc}}%
    \clrmacro%
    \textcolor{black}{#1}
}
\definecolor{c}{rgb}{1.0,0.67,0.0}
\def\pa#1{%
    \pgfmathsetmacro\calc{(#1-0.2)*100/(82.5-0.2)}
    \edef\clrmacro{\noexpand\cellcolor{c!\calc}}%
    \clrmacro%
    \textcolor{black}{#1}
}
\begin{table*}[]
\begin{center}
\newcolumntype{Y}{>{\centering\arraybackslash}X}
\scalebox{0.75}[0.75]{
\begin{tabularx}{\textwidth}{p{2cm}<{\centering}|YYYYYYYYYYYYYYYY}
 & \rotatebox{40}{DELF-FV} & \rotatebox{40}{DELF-VLAD}  & \rotatebox{40}{R-CroW} & \rotatebox{40}{R-GeM} & \rotatebox{40}{R-MAC} & \rotatebox{40}{R-rGeM} & \rotatebox{40}{R-rMAC} & \rotatebox{40}{R-SPoC} & \rotatebox{40}{V-CroW} & \rotatebox{40}{V-GeM} & \rotatebox{40}{V-MAC} & \rotatebox{40}{V-rGeM} & \rotatebox{40}{V-rMAC} & \rotatebox{40}{V-SPoC} & \rotatebox{40}{SIFT-FV} & \rotatebox{40}{SIFT-VLAD} \\ \hline \hline
DELF-FV& \ho{1.7} & \ho{4.4} & \ho{16.0} & \ho{19.3} & \ho{20.8} & \ho{17.9} & \ho{13.0} & \ho{16.9} & \ho{11.4} & \ho{14.1} & \ho{20.9} & \ho{13.0} & \ho{11.0} & \ho{13.6} & \ho{40.0} & \ho{42.5} \\
DELF-VLAD& \ho{4.0} & \ho{3.0} & \ho{15.9} & \ho{19.1} & \ho{21.3} & \ho{17.7} & \ho{14.0} & \ho{16.1} & \ho{10.5} & \ho{14.0} & \ho{21.0} & \ho{11.2} & \ho{11.3} & \ho{12.2} & \ho{40.7} & \ho{42.9} \\
R-CroW& \ho{12.9} & \ho{16.2} & \ho{1.2} & \ho{8.6} & \ho{4.9} & \ho{7.4} & \ho{3.2} & \ho{2.6} & \ho{8.8} & \ho{13.1} & \ho{17.9} & \ho{10.5} & \ho{8.2} & \ho{8.5} & \ho{32.6} & \ho{38.4} \\
R-GeM& \ho{10.2} & \ho{13.4} & \ho{6.4} & \ho{1.8} & \ho{5.5} & \ho{2.1} & \ho{3.0} & \ho{5.0} & \ho{6.7} & \ho{5.7} & \ho{12.6} & \ho{5.3} & \ho{7.1} & \ho{8.8} & \ho{33.5} & \ho{38.1} \\
R-MAC& \ho{12.7} & \ho{15.1} & \ho{2.8} & \ho{8.1} & \ho{4.1} & \ho{7.7} & \ho{1.8} & \ho{2.7} & \ho{6.2} & \ho{12.3} & \ho{8.8} & \ho{10.3} & \ho{4.6} & \ho{7.3} & \ho{38.5} & \ho{41.7} \\
R-rGeM& \ho{11.3} & \ho{12.8} & \ho{4.5} & \ho{2.1} & \ho{6.0} & \ho{1.7} & \ho{2.7} & \ho{5.5} & \ho{9.3} & \ho{6.7} & \ho{13.7} & \ho{4.7} & \ho{5.8} & \ho{9.9} & \ho{35.8} & \ho{40.0} \\
R-rMAC& \ho{11.6} & \ho{14.8} & \ho{1.8} & \ho{8.6} & \ho{4.3} & \ho{8.0} & \ho{2.0} & \ho{3.3} & \ho{7.6} & \ho{10.6} & \ho{11.6} & \ho{8.9} & \ho{5.2} & \ho{9.5} & \ho{37.2} & \ho{40.6} \\
R-SPoC& \ho{12.6} & \ho{15.7} & \ho{1.4} & \ho{9.1} & \ho{5.1} & \ho{8.0} & \ho{2.9} & \ho{2.6} & \ho{8.1} & \ho{13.0} & \ho{18.7} & \ho{11.0} & \ho{7.8} & \ho{8.2} & \ho{31.5} & \ho{36.7} \\
V-CroW& \ho{18.8} & \ho{20.0} & \ho{15.1} & \ho{17.7} & \ho{14.8} & \ho{18.4} & \ho{12.1} & \ho{15.3} & \ho{2.6} & \ho{9.8} & \ho{3.0} & \ho{9.8} & \ho{2.2} & \ho{3.8} & \ho{35.2} & \ho{38.1} \\
V-GeM& \ho{17.8} & \ho{19.6} & \ho{18.3} & \ho{14.0} & \ho{21.0} & \ho{15.2} & \ho{13.5} & \ho{20.1} & \ho{6.8} & \ho{3.5} & \ho{6.7} & \ho{2.8} & \ho{5.9} & \ho{9.8} & \ho{34.8} & \ho{38.4} \\
V-MAC& \ho{33.5} & \ho{36.7} & \ho{33.7} & \ho{34.6} & \ho{31.1} & \ho{35.3} & \ho{22.2} & \ho{35.8} & \ho{11.4} & \ho{18.9} & \ho{6.7} & \ho{20.9} & \ho{7.3} & \ho{15.2} & \ho{46.9} & \ho{50.5} \\
V-rGeM& \ho{18.0} & \ho{19.9} & \ho{17.2} & \ho{15.0} & \ho{20.2} & \ho{12.7} & \ho{12.4} & \ho{17.5} & \ho{8.9} & \ho{2.4} & \ho{9.9} & \ho{1.4} & \ho{5.8} & \ho{10.4} & \ho{35.4} & \ho{37.4} \\
V-rMAC& \ho{23.3} & \ho{26.1} & \ho{21.5} & \ho{25.9} & \ho{21.5} & \ho{23.3} & \ho{14.1} & \ho{22.9} & \ho{6.6} & \ho{12.8} & \ho{4.7} & \ho{12.6} & \ho{3.6} & \ho{9.9} & \ho{42.8} & \ho{45.1} \\
V-SPoC& \ho{17.2} & \ho{18.0} & \ho{13.6} & \ho{16.8} & \ho{14.7} & \ho{16.5} & \ho{11.1} & \ho{13.4} & \ho{1.8} & \ho{10.3} & \ho{5.7} & \ho{8.1} & \ho{3.6} & \ho{2.2} & \ho{30.9} & \ho{36.6} \\
SIFT-FV& \ho{55.9} & \ho{63.8} & \ho{61.2} & \ho{68.5} & \ho{69.5} & \ho{66.6} & \ho{57.1} & \ho{59.3} & \ho{59.5} & \ho{60.8} & \ho{63.4} & \ho{60.3} & \ho{59.4} & \ho{54.9} & \ho{3.7} & \ho{4.9} \\
SIFT-VLAD& \ho{57.9} & \ho{63.6} & \ho{61.4} & \ho{69.7} & \ho{70.7} & \ho{67.3} & \ho{56.0} & \ho{60.9} & \ho{60.5} & \ho{59.7} & \ho{64.8} & \ho{60.4} & \ho{60.4} & \ho{55.9} & \ho{1.6} & \ho{5.9} \\ \hline \hline

DELF-FV& \ox{4.8} & \ox{9.5} & \ox{15.5} & \ox{30.8} & \ox{22.2} & \ox{28.8} & \ox{18.5} & \ox{16.3} & \ox{11.7} & \ox{22.1} & \ox{18.4} & \ox{20.2} & \ox{21.2} & \ox{8.2} & \ox{30.3} & \ox{33.2} \\
DELF-VLAD& \ox{5.2} & \ox{4.2} & \ox{10.4} & \ox{27.0} & \ox{11.8} & \ox{25.5} & \ox{13.7} & \ox{9.7} & \ox{8.4} & \ox{19.6} & \ox{22.8} & \ox{17.4} & \ox{17.9} & \ox{7.9} & \ox{26.8} & \ox{30.1} \\
R-CroW& \ox{27.2} & \ox{27.2} & \ox{2.1} & \ox{24.4} & \ox{5.2} & \ox{21.3} & \ox{8.3} & \ox{2.8} & \ox{16.8} & \ox{27.1} & \ox{21.1} & \ox{21.3} & \ox{20.3} & \ox{15.3} & \ox{27.9} & \ox{30.6} \\
R-GeM& \ox{19.3} & \ox{15.8} & \ox{1.5} & \ox{2.6} & \ox{0.9} & \ox{3.1} & \ox{5.1} & \ox{3.4} & \ox{12.8} & \ox{11.3} & \ox{18.6} & \ox{11.2} & \ox{14.9} & \ox{12.5} & \ox{31.6} & \ox{33.9} \\
R-MAC& \ox{30.5} & \ox{28.0} & \ox{8.0} & \ox{26.4} & \ox{5.8} & \ox{25.8} & \ox{9.8} & \ox{6.9} & \ox{17.6} & \ox{27.9} & \ox{20.6} & \ox{26.8} & \ox{23.9} & \ox{18.3} & \ox{30.6} & \ox{33.3} \\
R-rGeM& \ox{17.8} & \ox{16.5} & \ox{1.4} & \ox{3.8} & \ox{1.1} & \ox{2.9} & \ox{4.8} & \ox{1.5} & \ox{11.4} & \ox{11.8} & \ox{19.2} & \ox{9.6} & \ox{14.6} & \ox{13.2} & \ox{29.7} & \ox{32.2} \\
R-rMAC& \ox{26.5} & \ox{24.6} & \ox{0.9} & \ox{21.0} & \ox{1.3} & \ox{19.1} & \ox{4.9} & \ox{1.5} & \ox{15.3} & \ox{22.4} & \ox{17.1} & \ox{18.9} & \ox{16.4} & \ox{15.3} & \ox{28.2} & \ox{31.8} \\
R-SPoC& \ox{25.9} & \ox{24.6} & \ox{2.1} & \ox{22.8} & \ox{4.4} & \ox{20.6} & \ox{7.0} & \ox{1.8} & \ox{15.3} & \ox{24.5} & \ox{21.7} & \ox{20.8} & \ox{19.5} & \ox{13.6} & \ox{27.0} & \ox{29.9} \\
V-CroW& \ox{23.5} & \ox{23.2} & \ox{13.6} & \ox{32.3} & \ox{19.0} & \ox{33.8} & \ox{14.6} & \ox{14.8} & \ox{1.0} & \ox{19.2} & \ox{0.3} & \ox{17.6} & \ox{6.2} & \ox{0.1} & \ox{26.7} & \ox{31.2} \\
V-GeM& \ox{17.1} & \ox{11.9} & \ox{11.3} & \ox{17.3} & \ox{15.9} & \ox{16.1} & \ox{9.1} & \ox{12.3} & \ox{3.2} & \ox{2.2} & \ox{4.7} & \ox{1.0} & \ox{6.5} & \ox{4.6} & \ox{29.3} & \ox{34.0} \\
V-MAC& \ox{40.0} & \ox{40.4} & \ox{33.2} & \ox{46.8} & \ox{29.1} & \ox{52.0} & \ox{30.2} & \ox{33.6} & \ox{9.9} & \ox{26.1} & \ox{5.4} & \ox{32.4} & \ox{10.5} & \ox{14.4} & \ox{30.9} & \ox{36.3} \\
V-rGeM& \ox{18.1} & \ox{13.4} & \ox{9.7} & \ox{21.6} & \ox{16.0} & \ox{17.0} & \ox{7.1} & \ox{10.8} & \ox{3.8} & \ox{4.1} & \ox{6.8} & \ox{1.9} & \ox{6.3} & \ox{7.1} & \ox{25.8} & \ox{29.7} \\
V-rMAC& \ox{31.3} & \ox{32.9} & \ox{21.4} & \ox{38.4} & \ox{20.3} & \ox{39.0} & \ox{18.4} & \ox{22.0} & \ox{3.7} & \ox{17.1} & \ox{0.9} & \ox{16.9} & \ox{1.3} & \ox{5.1} & \ox{27.6} & \ox{32.9} \\
V-SPoC& \ox{24.7} & \ox{22.5} & \ox{14.8} & \ox{38.3} & \ox{17.9} & \ox{36.4} & \ox{17.0} & \ox{16.0} & \ox{2.1} & \ox{19.1} & \ox{3.5} & \ox{17.3} & \ox{6.6} & \ox{0.5} & \ox{24.6} & \ox{30.6} \\
SIFT-FV& \ox{65.3} & \ox{67.9} & \ox{55.4} & \ox{80.9} & \ox{56.2} & \ox{79.5} & \ox{61.0} & \ox{57.2} & \ox{61.0} & \ox{77.1} & \ox{56.4} & \ox{75.3} & \ox{64.1} & \ox{59.2} & \ox{9.5} & \ox{13.5} \\
SIFT-VLAD& \ox{63.4} & \ox{67.3} & \ox{57.0} & \ox{81.2} & \ox{56.5} & \ox{79.7} & \ox{61.4} & \ox{57.2} & \ox{59.9} & \ox{76.6} & \ox{56.5} & \ox{75.2} & \ox{63.2} & \ox{57.6} & \ox{10.2} & \ox{9.8} \\ \hline \hline

DELF-FV& \pa{3.7} & \pa{6.0} & \pa{9.5} & \pa{20.1} & \pa{14.1} & \pa{18.8} & \pa{13.4} & \pa{13.4} & \pa{7.2} & \pa{13.4} & \pa{17.9} & \pa{13.5} & \pa{13.7} & \pa{6.3} & \pa{14.5} & \pa{16.5} \\
DELF-VLAD& \pa{6.4} & \pa{3.3} & \pa{8.9} & \pa{18.0} & \pa{13.0} & \pa{16.0} & \pa{11.5} & \pa{10.7} & \pa{5.5} & \pa{12.3} & \pa{19.1} & \pa{12.9} & \pa{13.7} & \pa{6.5} & \pa{15.0} & \pa{19.0} \\
R-CroW& \pa{16.8} & \pa{17.3} & \pa{4.0} & \pa{17.4} & \pa{6.0} & \pa{15.1} & \pa{8.9} & \pa{4.9} & \pa{11.0} & \pa{14.4} & \pa{18.4} & \pa{13.5} & \pa{14.9} & \pa{11.1} & \pa{17.6} & \pa{22.7} \\
R-GeM& \pa{10.7} & \pa{8.3} & \pa{0.2} & \pa{3.5} & \pa{1.9} & \pa{2.9} & \pa{0.6} & \pa{1.1} & \pa{2.7} & \pa{2.3} & \pa{5.7} & \pa{4.0} & \pa{7.6} & \pa{3.9} & \pa{19.4} & \pa{20.5} \\
R-MAC& \pa{18.9} & \pa{18.8} & \pa{4.7} & \pa{15.9} & \pa{7.9} & \pa{14.2} & \pa{9.0} & \pa{6.5} & \pa{12.3} & \pa{14.2} & \pa{18.0} & \pa{13.8} & \pa{18.1} & \pa{12.2} & \pa{20.6} & \pa{25.2} \\
R-rGeM& \pa{9.3} & \pa{9.4} & \pa{2.7} & \pa{3.5} & \pa{1.1} & \pa{3.2} & \pa{1.1} & \pa{0.4} & \pa{3.4} & \pa{3.7} & \pa{7.0} & \pa{3.9} & \pa{6.6} & \pa{3.8} & \pa{17.1} & \pa{20.7} \\
R-rMAC& \pa{14.2} & \pa{13.6} & \pa{0.2} & \pa{13.7} & \pa{3.4} & \pa{10.4} & \pa{4.9} & \pa{0.9} & \pa{6.9} & \pa{9.6} & \pa{11.9} & \pa{8.9} & \pa{9.8} & \pa{7.1} & \pa{18.2} & \pa{22.0} \\
R-SPoC& \pa{15.2} & \pa{15.2} & \pa{3.3} & \pa{17.1} & \pa{5.1} & \pa{14.3} & \pa{8.0} & \pa{3.6} & \pa{10.0} & \pa{13.6} & \pa{15.8} & \pa{12.7} & \pa{13.2} & \pa{9.7} & \pa{16.8} & \pa{22.0} \\
V-CroW& \pa{18.1} & \pa{20.0} & \pa{10.4} & \pa{22.9} & \pa{13.9} & \pa{23.2} & \pa{13.9} & \pa{13.5} & \pa{1.0} & \pa{10.9} & \pa{1.7} & \pa{9.6} & \pa{5.0} & \pa{0.3} & \pa{19.3} & \pa{21.6} \\
V-GeM& \pa{10.6} & \pa{12.2} & \pa{7.4} & \pa{11.4} & \pa{8.0} & \pa{11.3} & \pa{6.7} & \pa{10.0} & \pa{1.8} & \pa{1.9} & \pa{1.4} & \pa{2.0} & \pa{4.8} & \pa{2.3} & \pa{13.5} & \pa{17.7} \\
V-MAC& \pa{29.6} & \pa{33.0} & \pa{24.9} & \pa{31.4} & \pa{24.7} & \pa{34.6} & \pa{23.6} & \pa{29.3} & \pa{8.7} & \pa{15.1} & \pa{7.3} & \pa{16.3} & \pa{9.7} & \pa{9.2} & \pa{26.8} & \pa{30.4} \\
V-rGeM& \pa{10.9} & \pa{12.8} & \pa{6.2} & \pa{12.3} & \pa{6.8} & \pa{12.3} & \pa{5.8} & \pa{6.5} & \pa{1.2} & \pa{3.3} & \pa{3.1} & \pa{1.2} & \pa{4.8} & \pa{1.7} & \pa{12.6} & \pa{16.0} \\
V-rMAC& \pa{21.4} & \pa{24.2} & \pa{12.9} & \pa{25.0} & \pa{19.5} & \pa{22.3} & \pa{15.4} & \pa{14.9} & \pa{1.7} & \pa{9.7} & \pa{1.0} & \pa{8.5} & \pa{2.7} & \pa{1.8} & \pa{20.0} & \pa{24.2} \\
V-SPoC& \pa{16.9} & \pa{21.8} & \pa{11.6} & \pa{23.8} & \pa{14.2} & \pa{25.6} & \pa{14.1} & \pa{13.8} & \pa{2.5} & \pa{13.2} & \pa{2.7} & \pa{12.7} & \pa{6.5} & \pa{1.6} & \pa{15.4} & \pa{19.7} \\
SIFT-FV& \pa{59.8} & \pa{64.0} & \pa{59.3} & \pa{82.2} & \pa{67.3} & \pa{78.9} & \pa{62.3} & \pa{61.7} & \pa{63.0} & \pa{71.6} & \pa{63.0} & \pa{68.9} & \pa{66.9} & \pa{60.1} & \pa{8.8} & \pa{10.0} \\
SIFT-VLAD& \pa{58.7} & \pa{60.7} & \pa{60.7} & \pa{80.5} & \pa{67.4} & \pa{78.2} & \pa{63.5} & \pa{61.4} & \pa{62.3} & \pa{70.9} & \pa{63.6} & \pa{68.7} & \pa{65.8} & \pa{59.1} & \pa{5.9} & \pa{10.2} \\ \hline
\end{tabularx}}
\end{center}
\caption{The mAP(\%) difference between target and translated features on three public datasets: Holidays (\textcolor[rgb]{0.6,0.8,0.5}{Green}), Oxford5k (\textcolor[rgb]{0.7,0.8,0.9}{Blue}) and Paris6k (\textcolor[rgb]{1.0,0.67,0.0}{Brown}) in the first, second and third blocks, respectively.}
\label{table2}
\end{table*}
\begin{figure*}[t]
\centering
\includegraphics[width=6.8in]{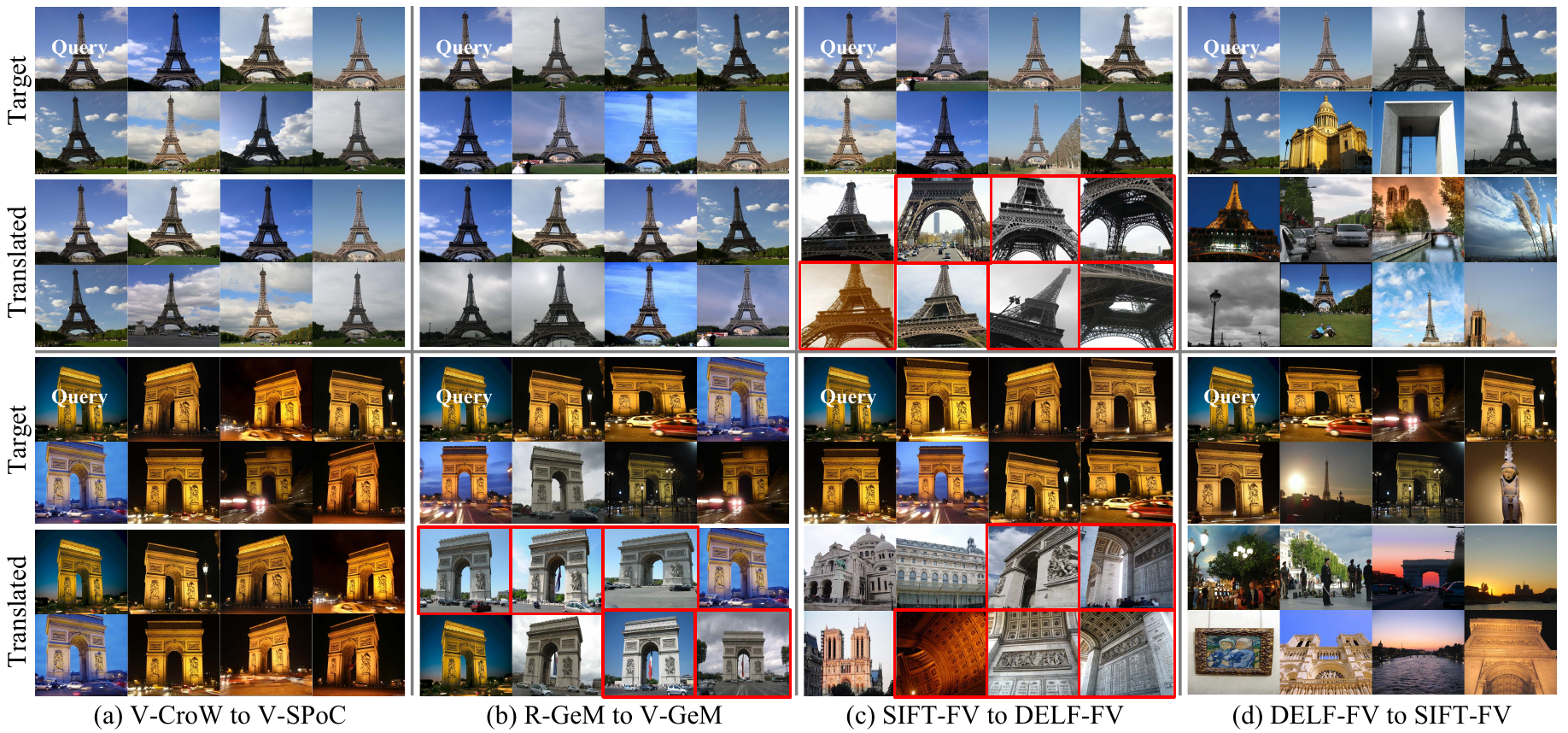}
\caption{
The retrieval results for querying images of the \textit{Eiffel Tower} (up) and the \textit{Arc de Triomphe} (down) with the target features and the translated features.
The images are resized for better view and the interesting results are colored by red bounding boxes.
}
\label{fig4}
\end{figure*}
\definecolor{a}{rgb}{0.6,0.8,0.5}
\def\mlp#1{%
    \pgfmathsetmacro\calc{(#1-0.0767)*100/(87.0167-0.0767))}
    \edef\clrmacro{\noexpand\cellcolor{a!\calc}}%
    \clrmacro%
    \textcolor{black}{#1}
}
\definecolor{b}{rgb}{0.7,0.8,0.9}
\def\hae#1{%
    \pgfmathsetmacro\calc{(#1-0.9700)*100/(77.1833-0.9700)}
    \edef\clrmacro{\noexpand\cellcolor{b!\calc}}%
    \clrmacro%
    \textcolor{black}{#1}
}
\begin{table}[]
\begin{center}
\newcolumntype{Y}{>{\centering\arraybackslash}X}
\scalebox{0.43}[0.43]{
\begin{tabularx}{\textwidth}{p{2cm}<{\centering}|YYYYYYYYYYYYYYYY}
 & \rotatebox{50}{DELF-FV} & \rotatebox{50}{DELF-VLAD}  & \rotatebox{50}{R-CroW} & \rotatebox{50}{R-GeM} & \rotatebox{50}{R-MAC} & \rotatebox{50}{R-rGeM} & \rotatebox{50}{R-rMAC} & \rotatebox{50}{R-SPoC} & \rotatebox{50}{V-CroW} & \rotatebox{50}{V-GeM} & \rotatebox{50}{V-MAC} & \rotatebox{50}{V-rGeM} & \rotatebox{50}{V-rMAC} & \rotatebox{50}{V-SPoC} & \rotatebox{50}{SIFT-FV} & \rotatebox{50}{SIFT-VLAD} \\ \hline
DELF-FV& \mlp{1.5} & \mlp{5.9} & \mlp{72.0} & \mlp{85.8} & \mlp{73.6} & \mlp{86.0} & \mlp{77.9} & \mlp{72.7} & \mlp{9.8} & \mlp{17.4} & \mlp{18.5} & \mlp{15.3} & \mlp{16.1} & \mlp{9.7} & \mlp{43.1} & \mlp{46.9} \\
DELF-VLAD& \mlp{4.8} & \mlp{1.7} & \mlp{71.8} & \mlp{85.7} & \mlp{73.1} & \mlp{86.0} & \mlp{77.4} & \mlp{72.4} & \mlp{9.4} & \mlp{14.5} & \mlp{20.1} & \mlp{13.9} & \mlp{16.9} & \mlp{9.0} & \mlp{43.3} & \mlp{47.1} \\
R-CroW& \mlp{76.7} & \mlp{77.3} & \mlp{1.1} & \mlp{83.3} & \mlp{3.5} & \mlp{15.2} & \mlp{4.6} & \mlp{1.6} & \mlp{11.2} & \mlp{18.3} & \mlp{19.7} & \mlp{14.9} & \mlp{13.6} & \mlp{10.9} & \mlp{42.8} & \mlp{47.0} \\
R-GeM& \mlp{76.9} & \mlp{77.2} & \mlp{71.2} & \mlp{1.0} & \mlp{58.8} & \mlp{0.9} & \mlp{9.4} & \mlp{72.7} & \mlp{7.0} & \mlp{6.6} & \mlp{12.7} & \mlp{5.9} & \mlp{9.7} & \mlp{7.0} & \mlp{43.2} & \mlp{47.0} \\
R-MAC& \mlp{77.0} & \mlp{77.5} & \mlp{2.9} & \mlp{65.6} & \mlp{2.3} & \mlp{84.9} & \mlp{4.0} & \mlp{3.6} & \mlp{12.7} & \mlp{18.6} & \mlp{15.8} & \mlp{16.9} & \mlp{14.8} & \mlp{12.5} & \mlp{43.2} & \mlp{47.2} \\
R-rGeM& \mlp{77.7} & \mlp{76.8} & \mlp{4.0} & \mlp{1.2} & \mlp{69.4} & \mlp{0.9} & \mlp{3.2} & \mlp{22.9} & \mlp{7.5} & \mlp{7.3} & \mlp{15.3} & \mlp{5.4} & \mlp{9.2} & \mlp{7.7} & \mlp{43.2} & \mlp{46.8} \\
R-rMAC& \mlp{76.1} & \mlp{76.7} & \mlp{0.6} & \mlp{25.7} & \mlp{0.1} & \mlp{10.5} & \mlp{1.1} & \mlp{0.2} & \mlp{9.9} & \mlp{15.7} & \mlp{14.1} & \mlp{12.3} & \mlp{10.0} & \mlp{9.4} & \mlp{43.1} & \mlp{47.0} \\
R-SPoC& \mlp{76.8} & \mlp{77.3} & \mlp{0.7} & \mlp{82.0} & \mlp{3.0} & \mlp{15.2} & \mlp{4.2} & \mlp{1.4} & \mlp{10.3} & \mlp{17.3} & \mlp{18.8} & \mlp{14.1} & \mlp{12.3} & \mlp{9.7} & \mlp{42.9} & \mlp{46.9} \\
V-CroW& \mlp{21.3} & \mlp{25.1} & \mlp{14.8} & \mlp{24.7} & \mlp{17.1} & \mlp{27.8} & \mlp{15.5} & \mlp{15.0} & \mlp{0.9} & \mlp{11.1} & \mlp{0.1} & \mlp{10.0} & \mlp{3.2} & \mlp{0.3} & \mlp{43.1} & \mlp{46.5} \\
V-GeM& \mlp{16.6} & \mlp{23.2} & \mlp{18.7} & \mlp{15.3} & \mlp{27.7} & \mlp{15.2} & \mlp{12.6} & \mlp{22.0} & \mlp{1.8} & \mlp{1.2} & \mlp{0.2} & \mlp{0.7} & \mlp{3.4} & \mlp{2.8} & \mlp{43.1} & \mlp{46.2} \\
V-MAC& \mlp{44.6} & \mlp{66.4} & \mlp{53.1} & \mlp{71.8} & \mlp{51.5} & \mlp{69.9} & \mlp{37.0} & \mlp{62.7} & \mlp{8.7} & \mlp{16.4} & \mlp{3.5} & \mlp{17.9} & \mlp{6.9} & \mlp{10.6} & \mlp{43.3} & \mlp{46.9} \\
V-rGeM& \mlp{20.7} & \mlp{24.3} & \mlp{14.2} & \mlp{18.7} & \mlp{20.3} & \mlp{13.6} & \mlp{10.3} & \mlp{12.9} & \mlp{3.0} & \mlp{2.6} & \mlp{3.5} & \mlp{1.1} & \mlp{3.4} & \mlp{3.2} & \mlp{43.0} & \mlp{46.3} \\
V-rMAC& \mlp{29.3} & \mlp{42.1} & \mlp{22.9} & \mlp{34.1} & \mlp{24.3} & \mlp{32.7} & \mlp{17.8} & \mlp{22.3} & \mlp{2.8} & \mlp{11.9} & \mlp{1.3} & \mlp{10.2} & \mlp{1.5} & \mlp{3.6} & \mlp{42.2} & \mlp{46.1} \\
V-SPoC& \mlp{19.6} & \mlp{22.6} & \mlp{13.3} & \mlp{26.9} & \mlp{16.3} & \mlp{25.3} & \mlp{15.8} & \mlp{14.4} & \mlp{1.9} & \mlp{12.5} & \mlp{2.0} & \mlp{10.7} & \mlp{4.9} & \mlp{0.9} & \mlp{41.1} & \mlp{45.8} \\
SIFT-FV& \mlp{78.3} & \mlp{79.1} & \mlp{72.9} & \mlp{87.0} & \mlp{74.2} & \mlp{87.0} & \mlp{78.6} & \mlp{73.8} & \mlp{59.8} & \mlp{74.6} & \mlp{67.8} & \mlp{77.1} & \mlp{76.2} & \mlp{57.3} & \mlp{4.1} & \mlp{12.6} \\
SIFT-VLAD& \mlp{78.0} & \mlp{78.8} & \mlp{72.9} & \mlp{87.0} & \mlp{73.9} & \mlp{86.8} & \mlp{78.5} & \mlp{73.6} & \mlp{59.6} & \mlp{77.1} & \mlp{67.9} & \mlp{80.6} & \mlp{77.6} & \mlp{57.9} & \mlp{8.6} & \mlp{4.3} \\ \hline
DELF-FV& \hae{3.4} & \hae{6.6} & \hae{13.7} & \hae{23.4} & \hae{19.0} & \hae{21.8} & \hae{15.0} & \hae{15.5} & \hae{10.1} & \hae{16.5} & \hae{19.1} & \hae{15.6} & \hae{15.3} & \hae{9.4} & \hae{28.2} & \hae{30.7} \\
DELF-VLAD& \hae{5.2} & \hae{3.5} & \hae{11.7} & \hae{21.4} & \hae{15.4} & \hae{19.7} & \hae{13.1} & \hae{12.2} & \hae{8.1} & \hae{15.3} & \hae{21.0} & \hae{13.8} & \hae{14.3} & \hae{8.9} & \hae{27.5} & \hae{30.7} \\
R-CroW& \hae{18.9} & \hae{20.2} & \hae{2.4} & \hae{16.8} & \hae{5.4} & \hae{14.6} & \hae{6.8} & \hae{3.4} & \hae{12.2} & \hae{18.2} & \hae{19.1} & \hae{15.1} & \hae{14.5} & \hae{11.7} & \hae{26.0} & \hae{30.6} \\
R-GeM& \hae{13.4} & \hae{12.5} & \hae{2.6} & \hae{2.6} & \hae{1.5} & \hae{2.7} & \hae{2.9} & \hae{3.1} & \hae{7.4} & \hae{6.4} & \hae{12.3} & \hae{6.9} & \hae{9.8} & \hae{8.4} & \hae{28.2} & \hae{30.8} \\
R-MAC& \hae{20.7} & \hae{20.6} & \hae{5.2} & \hae{16.8} & \hae{5.9} & \hae{15.9} & \hae{6.8} & \hae{5.4} & \hae{12.0} & \hae{18.1} & \hae{15.8} & \hae{16.9} & \hae{15.5} & \hae{12.6} & \hae{29.9} & \hae{33.4} \\
R-rGeM& \hae{12.8} & \hae{12.9} & \hae{1.1} & \hae{3.2} & \hae{2.0} & \hae{2.6} & \hae{2.9} & \hae{2.2} & \hae{8.0} & \hae{7.4} & \hae{13.3} & \hae{6.1} & \hae{9.0} & \hae{9.0} & \hae{27.6} & \hae{31.0} \\
R-rMAC& \hae{17.4} & \hae{17.7} & \hae{1.0} & \hae{14.4} & \hae{3.0} & \hae{12.5} & \hae{3.9} & \hae{1.9} & \hae{9.9} & \hae{14.2} & \hae{13.5} & \hae{12.3} & \hae{10.5} & \hae{10.6} & \hae{27.9} & \hae{31.5} \\
R-SPoC& \hae{17.9} & \hae{18.5} & \hae{2.3} & \hae{16.3} & \hae{4.9} & \hae{14.3} & \hae{6.0} & \hae{2.7} & \hae{11.1} & \hae{17.1} & \hae{18.7} & \hae{14.9} & \hae{13.5} & \hae{10.5} & \hae{25.1} & \hae{29.5} \\
V-CroW& \hae{20.2} & \hae{21.1} & \hae{13.0} & \hae{24.3} & \hae{15.9} & \hae{25.1} & \hae{13.5} & \hae{14.5} & \hae{1.6} & \hae{13.3} & \hae{1.7} & \hae{12.3} & \hae{4.5} & \hae{1.4} & \hae{27.1} & \hae{30.3} \\
V-GeM& \hae{15.2} & \hae{14.6} & \hae{12.3} & \hae{14.2} & \hae{15.0} & \hae{14.2} & \hae{9.8} & \hae{14.2} & \hae{3.9} & \hae{2.6} & \hae{4.3} & \hae{1.9} & \hae{5.7} & \hae{5.6} & \hae{25.9} & \hae{30.0} \\
V-MAC& \hae{34.3} & \hae{36.7} & \hae{30.6} & \hae{37.6} & \hae{28.3} & \hae{40.6} & \hae{25.3} & \hae{32.9} & \hae{10.0} & \hae{20.0} & \hae{6.5} & \hae{23.2} & \hae{9.2} & \hae{12.9} & \hae{34.9} & \hae{39.1} \\
V-rGeM& \hae{15.7} & \hae{15.4} & \hae{11.0} & \hae{16.3} & \hae{14.3} & \hae{14.0} & \hae{8.4} & \hae{11.6} & \hae{4.6} & \hae{3.3} & \hae{6.6} & \hae{1.5} & \hae{5.6} & \hae{6.4} & \hae{24.6} & \hae{27.7} \\
V-rMAC& \hae{25.4} & \hae{27.7} & \hae{18.6} & \hae{29.7} & \hae{20.4} & \hae{28.2} & \hae{15.9} & \hae{20.0} & \hae{4.0} & \hae{13.2} & \hae{2.2} & \hae{12.7} & \hae{2.5} & \hae{5.6} & \hae{30.2} & \hae{34.1} \\
V-SPoC& \hae{19.6} & \hae{20.8} & \hae{13.4} & \hae{26.3} & \hae{15.6} & \hae{26.2} & \hae{14.1} & \hae{14.4} & \hae{2.1} & \hae{14.2} & \hae{4.0} & \hae{12.7} & \hae{5.6} & \hae{1.4} & \hae{23.6} & \hae{29.0} \\
SIFT-FV& \hae{60.3} & \hae{65.2} & \hae{58.7} & \hae{77.2} & \hae{64.3} & \hae{75.0} & \hae{60.1} & \hae{59.4} & \hae{61.2} & \hae{69.8} & \hae{60.9} & \hae{68.2} & \hae{63.5} & \hae{58.1} & \hae{7.3} & \hae{9.5} \\
SIFT-VLAD& \hae{60.0} & \hae{63.9} & \hae{59.7} & \hae{77.1} & \hae{64.9} & \hae{75.1} & \hae{60.3} & \hae{59.8} & \hae{60.9} & \hae{69.1} & \hae{61.6} & \hae{68.1} & \hae{63.1} & \hae{57.5} & \hae{5.9} & \hae{8.6} \\ \hline
\end{tabularx}}
\end{center}
\caption{The average mAP difference (\%) of MLP (\textcolor[rgb]{0.6,0.8,0.5}{Green}) and HAE (\textcolor[rgb]{0.7,0.8,0.9}{Blue}) on three datasets.}
\label{table3}
\end{table}
\subsection{Translation Results}
\label{sec:Translation_result}
\textbf{Quantitative Evaluation.} The performance of target features is shown in Table~\ref{table1}.
We use the mAP difference between target and translated features to show the translation results.
As shown in Table~\ref{table2}, we use a color map which is normalized according to the minimum (white) and maximum (colored) values to show results of each dataset.
From the result, we find although there are still few differences between datasets, the trend of the colored values is almost the same.
For further analyzing, the results can be divided into three groups: high convertibility, inferior convertibility and low convertibility.
Firstly, the high convertibility results appear mostly in the translation between homologous features.
For example, when translating from V-CroW to V-SPoC, the mAPs drop 3.8, 0.1, 0.3 on the Holidays, Oxford5k and Paris6k datasets, respectively.
Secondly, the inferior results are found between heterogenous features such as R-based features and V-based features.
For example, when translating from R-GeM to V-GeM, the mAPs decrease 5.7, 11.3, 2.3 on the three datasets, respectively.
Another example is the translation from V-rGeM to R-rMAC, the mAPs decrease 12.4, 7.1, 5.8 on the three datasets, respectively.
Thirdly, the low convertibility results also emerge between heterogenous features.
For example, when translating from SIFT-FV to DELF-FV, the performance is not high.
Another example is the translation from DELF-VLAD to R-GeM, in which the former is extracted by Resnet50 and the latter is extracted by Resnet101.
We explain it from the different depth of network architectures, different training procedures and different encoding/pooling schemes.
The average mAP difference of HAE compared with MLP on three datasets is shown in Table~\ref{table3}.
From the results, we can see MLP has a very unstable performance.
In contrast, HAE with appropriate dimension of latent feature performs better than MLP, due to the regularization effect brought by the ``bottleneck" architecture (which enforces encoder to learn the most valuable information for decoder).
\textbf{Qualitative Evaluation.} Some cross-feature retrieval results are shown in Fig.~\ref{fig4}.
The first column shows a successful translation from V-CroW to V-SPoC, the ranking lists are almost the same.
The second column shows an inferior translation from R-GeM to V-GeM.
Interestingly, when querying an image of the \textit{Arc de Triomphe} at night, the images of the \textit{Arc de Triomphe} during the day are retrieved by the translated features and get high ranks, which inspires the integration of feature translation to improve cross-modal retrieval.
The most exciting result lies in the third column: although the translation from SIFT-FV to DELF-FV suffers a low performance, the characteristics like rotation or viewpoint invariance can be highlighted by translation, which well bridges the merits of the handcrafted features to the learning-based features.
For example, the images from the bottom view of the \textit{Eiffel Tower} and the \textit{Arc de Triomphe} get high ranks (both at Rank@4).
The rotated images of them also have high ranks (at Rank@7 and Rank@3).
Then, in the fourth column, we show these characteristics do not symmetrically exist in the reverse translation from DELF-FV to SIFT-FV.
We explain it from the limited representative ability of the SIFT-FV.
\subsection{Relation Mining Results}
\label{sec:Relation_result}
\begin{figure}[t]
\centering
\includegraphics[width=3.25in]{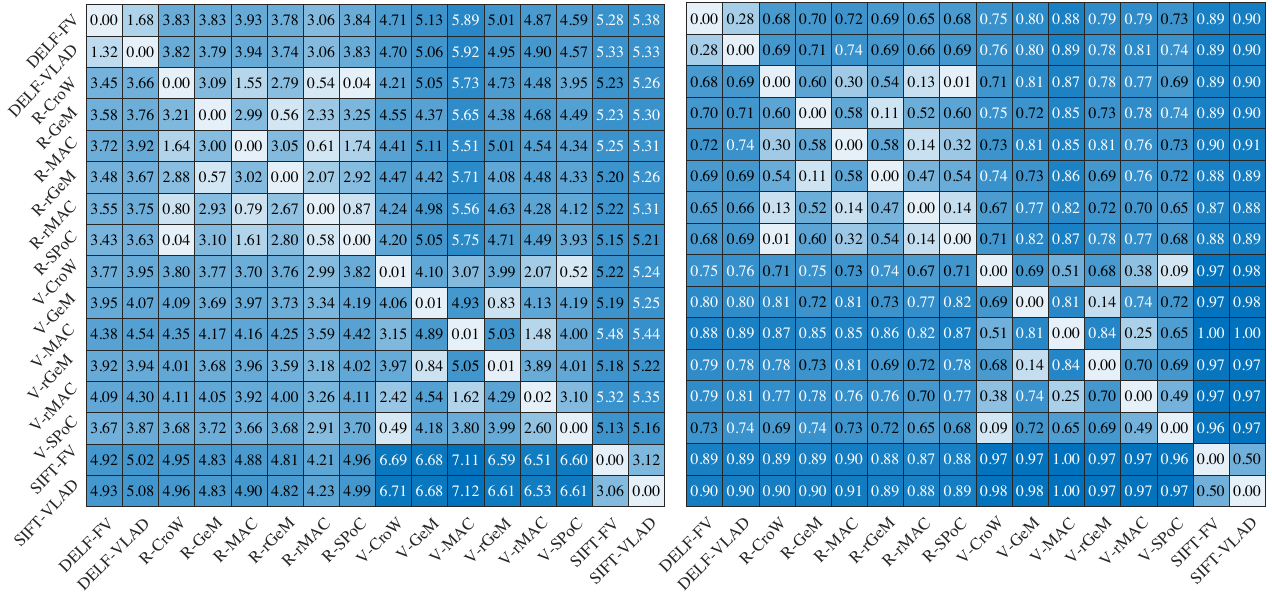}
\caption{
The heat maps of the directed affinity matrix $M$ (left) and the undirected affinity matrix $U$ (right), the values are the averaged results on Holidays, Oxford5k and Paris6k datasets.
}
\label{fig5}
\end{figure}

After calculating the directed affinity matrix $M$ and the undirected affinity matrix $U$, we average the values of the three datasets and draw the heat maps.
As shown in Fig.~\ref{fig5} (left), the values of directed affinity matrix $M$ verify our assumption that the reconstruction error is smaller than the translation error as all the values are positive.
As shown in Fig.~\ref{fig5} (right), the positions of light and dark colors are almost the same as that of the translation results in Table~\ref{table2}, which indicates the UAM can be used to predict the translation quality between two given features.
To study the relationship between features better, we visualize the MST based on $U$ as Fig.~\ref{fig3}.
The images are the ranking lists for a query image with corresponding features.
Since the results of leaf nodes connected in the MST (\eg R-CroW and R-SPoC) are very similar, we mainly show the results of nodes in the trunk of the MST.
The closer features return more similar ranking lists, which indicates the rationality of our affinity measurement from the other perspective.
\section{Conclusion}
\label{sec:conclusion}
In this work, we present the first attempt to investigate visual feature translation, as well as the first attempt at quantifying the affinity among different types of features in visual search.
In particular, we propose a Hybrid Auto-Encoder (HAE) to translate visual features.
Based on HAE, we design an Undirected Affinity Measurement (UAM) to quantify the affinity.
Extensive experiments have been conducted on several public datasets with $16$ different types of widely-used features in visual search.
Quantitative results prove the encouraging possibility of feature translation.

\section*{Acknowledgments}
This work is supported by the National Key R\&D Program (No.2017YFC0113000, and No.2016YFB1001503), the Natural Science Foundation of China (No.U1705262, No.61772443, No.61402388 and No.61572410), the Post Doctoral Innovative Talent Support Program under Grant BX201600094, the China Post-Doctoral Science Foundation under Grant 2017M612134, Scientific Research Project of National Language Committee of China (Grant No. YB135-49), and Natural Science Foundation of Fujian Province, China (No. 2017J01125 and No. 2018J01106).

{\small
\bibliographystyle{ieee_fullname}
\bibliography{egbib}
}

\end{document}